\documentclass{Interspeech2024}
\usepackage{multirow}
\usepackage{caption}
\usepackage{booktabs}
\newcolumntype{C}[1]{>{\centering\arraybackslash}p{#1}}

\interspeechcameraready

\title{Weighted Cross-entropy for Low-Resource Languages in\\Multilingual Speech Recognition\vspace{-2ex}}

\name[affiliation={1,2}]{Andrés}{Piñeiro-Martín}
\name[affiliation={1}]{Carmen}{García-Mateo}
\name[affiliation={1}]{Laura}{Docío-Fernández}
\name[affiliation={2}]{\\María del Carmen}{López-Pérez}
\name[affiliation={3}]{Georg}{Rehm}

\address{
\fontsize{11}{12}\selectfont
  $^1$GTM research group, AtlanTTic Research Center, University of Vigo, Vigo, Spain\\
  $^2$Balidea Consulting \& Programming S.L., Santiago de Compostela, Spain\\
  $^3$DFKI GmbH, Speech and Language Technology Lab, Berlin, Germany
  \normalsize}
\email{andres.pinerio@balidea.com}


\begin{document}

\maketitle

\begin{abstract}
    
    This paper addresses the challenge of integrating low-resource languages into multilingual automatic speech recognition (ASR) systems. We introduce a novel application of weighted cross-entropy, typically used for unbalanced datasets, to facilitate the integration of low-resource languages into pre-trained multilingual ASR models within the context of continual multilingual learning. We fine-tune the Whisper multilingual ASR model on five high-resource languages and one low-resource language, employing language-weighted dynamic cross-entropy and data augmentation. The results show a remarkable 6.69\% word error rate (WER) reduction for the low-resource language compared to the fine-tuned model without applying our approach, and a 48.86\% WER reduction compared to the original Whisper model. In addition, our approach yields an average WER reduction of 3.29\% across the six languages, showing no degradation for the high-resource languages. 
\end{abstract}

\section{Introduction}

In recent years, significant advances in multilingual automatic speech recognition (ASR) models have revolutionised the language technology landscape. These breakthroughs have enabled the development of ASR systems capable of understanding and transcribing speech from an increasingly diverse range of languages, including many low-resource languages. This progress has been made possible by combining multilingual pre-trained ASR models with fine-tuning techniques \cite{baevski2020effectiveness} that require only a few hours of labelled speech data to extend ASR capabilities to low-resource languages.

Despite the impressive results for low-resource languages, their performance still lags behind that of high-resource languages, especially when using single multilingual models. The integration of low-resource and under-represented languages into pre-trained multilingual models aims to achieve comparable performance to high-resource languages without compromising overall results. However, this objective poses a major challenge due to unbalanced datasets resulting from data scarcity during pre-training and fine-tuning, making it a hot topic in research \cite{zhao2022improving,miao2022multilingual}.

Continual multilingual learning \cite{hadsell2020embracing} is a powerful paradigm that allows models to learn from data in multiple languages without forgetting previously acquired ``knowledge''. As the model encounters new languages with limited data, adapting to their linguistic features while retaining knowledge of previous languages becomes a critical challenge. Researchers have explored several techniques to effectively address this issue. Elastic weight consolidation (EWC) \cite{kirkpatrick2017overcoming} helps to mitigate catastrophic forgetting, while rehearsal or replay methods allow for periodic revisiting of previously seen data to maintain performance on earlier languages. Transfer learning \cite{wang2015transfer} has proved valuable for exploiting knowledge from high-resource languages, and data augmentation techniques improve the model's ability to generalise to different speech patterns.

In this paper, we propose a novel and straightforward approach of continual learning to improve the performance of multilingual ASR models for low-resource languages\footnote{\urlstyle{same}\url{https://github.com/andrespimartin/weighted-x-entropy-asr}}
. Our research focuses on applying language-weighted dynamic cross-entropy and data augmentation, not to deal with unbalanced datasets, but with unbalanced and biased pre-trained models. By applying weighted cross-entropy, we assign greater importance to low-resource languages during training, and we explore evolving weight approaches in cross-entropy calculations to identify the most effective strategy. In addition, the simultaneous application of data augmentation on the target language emerges as a technique to smooth the application of weighted cross-entropy and prevent degradation in the other languages.

We conduct our experiments using the well-known Whisper model \cite{radford2023robust}, a weakly supervised pre-trained multilingual ASR model that uses a large amount of labelled data. We fine-tune the model on five high-resource languages -- Spanish, Portuguese, French, German, and English -- and one low-resource language -- Galician.
For training, we use the Common Voice dataset \cite{ardila2020common}, employing all available resources for Galician, and ensuring balanced data for the other languages. Since the test split is used for training in Galician, and in order to test the robustness of the multilingual model for a domain different from the training domain and against a larger and more diverse dataset, we use the FalAI dataset \cite{pineiro2024falai} for testing Galician.

The results show significant improvements in recognition accuracy for the low-resource language, as evidenced by a remarkable 6.69\% reduction in Word Error Rate (WER) compared to the fine-tuned model without weighted cross-entropy, and a 48.86\% relative reduction in WER compared to the original Whisper model.
On average, our approach achieves a 3.29\% WER relative reduction across the six languages under study when compared to the simple fine-tuned model, with no degradation for the high-resource languages. When compared to the original Whisper model, our approach showcases a substantial average WER reduction of 32.5\%.

The rest of the paper is structured as follows: Section~\ref{sec:rel_work} outlines related work on low-resource languages in ASR. Section~\ref{sec:wce} details our approaches to weighted cross-entropy computation. Sections~\ref{sec:exp_setup} and~\ref{sec:results} cover the experimental setup and results. Conclusions and future directions are given in Section~\ref{sec:conclusions}.

\section{Low-resource languages in ASR}
\label{sec:rel_work}

The emergence of large pre-trained multilingual models has opened up opportunities for low-resource languages to achieve levels of performance previously unattainable with limited labelled data. State-of-the-art pre-trained models have mainly followed two main approaches to exploiting the potential of multilingual transfer learning.

The first approach involves self-supervised learning, where models leverage large amounts of multilingual unlabelled data to learn cross-lingual speech representations, enabling them to capture linguistic patterns and relationships across multiple languages \cite{baevski2019vq}. This approach has been used in several prominent multilingual models, such as the XLS-R model \cite{babu22_interspeech}, the Universial Speech Model \cite{zhang2023google}, the Massively Multilingual Speech project \cite{pratap2023scaling}, which has demonstrated scalability to 1,000 languages, and the more recent Wav2Vec-BERT 2.0, introduced in the family of Seamless Communication models \cite{barrault2023seamless}.

The second approach is based on weakly supervised pre-trained models, represented by Whisper \cite{radford2023robust}, which has been trained on a large dataset of 680k hours of recordings. This leap in training data volume represents a significant shift in model capabilities and performance. These models have demonstrated increased robustness and superior generalisation to previously unseen datasets \cite{narayanan2018toward,likhomanenko21_interspeech}, and can be used without fine-tuning.

However, the results for low-resource languages are still not comparable to those for high-resource languages, mainly due to unbalanced datasets, which leads to biased model predictions and reduced performance for low-resource languages. To address this issue, researchers have proposed different strategies to mitigate the impact of data imbalance, such as transfer learning \cite{khare2021low} or data augmentation \cite{meng2021mixspeech}. Data augmentation has proven to be instrumental in enhancing the performance of low-resource ASR models \cite{ragni2014data}. By introducing diverse variations in the speech data, such as different speaking rates, background noise, and acoustic conditions, data augmentation helps the model to become more robust and better able to generalise across different scenarios.

Another effective strategy to mitigate data imbalance is the application of weighted cross-entropy in the calculation of the loss function. This method involves assigning higher weights to underrepresented classes, i.e., increasing their influence during model training. Weighted cross-entropy has been successfully applied to various challenges in classification \cite{aurelio2019learning}, object detection \cite{chen2017learning} and keyword spotting \cite{panchapagesan16_interspeech}. However, the impact of this strategy has primarily been investigated in the context of correcting imbalances in training datasets. Research exploring how effectively it addresses imbalances in pre-trained models during continuous multilingual learning remains scarce, particularly in the area of multilingual and multitask ASR models.

In this work, we propose the combination of data augmentation together with weighted cross-entropy to address the incorporation of a low-resource language into a multilingual ASR model. Our work delves into the investigation of techniques for dealing with low-resource scenarios and overcoming the challenge of an imbalanced pre-trained model. We propose the use of language-weighted dynamic cross-entropy together with simple data augmentation techniques to incorporate Galician, generally considered a low-resource language \cite{Gaspari2023} with poor performance in the original Whisper model, into the multilingual model through fine-tuning. In our study, we explore the effect of weights and weight progressions for cross-entropy on recognition results.

\section{Weighted Cross-Entropy}
\label{sec:wce}

Whisper is pre-trained and fine-tuned to correctly classify the target text token from a predefined vocabulary of text tokens, using cross-entropy as the loss function. This is a standard objective function for training sequence-to-sequence systems on classification tasks, and it measures the discrepancy between the predicted probability distributions by the model and labels. The cross-entropy loss \( \mathcal{L}(y, p) \) of a single sentence is formulated as:

\begin{equation}
\mathcal{L}(y, p) = - \sum_{i=1}^{N} y_i \cdot \log(p_i)
\end{equation}

where \( N \) represents the total number of tokens (we use a multilingual tokenizer with the same number of tokens for each language), \( y_i \) represents the ground truth label corresponding to class \( i \), and \( p_i \) the probabilities predicted by the system for class \( i \), which are the model's outputs before applying the softmax function.

In addition to standard cross-entropy, a variant known as weighted cross-entropy can be used to further modulate the loss function according to the importance of different classes, or in our case, different languages. Weighted cross-entropy introduces weights for each class to reflect its importance in the learning process. For each batch, we define a weight vector \( \mathbf{w} \in \mathbb{R}^K \) with elements \( w_k > 0 \) defined over the range of language labels \( k \in \{1, 2, ..., K\} \). We define the \emph{language-weighted cross-entropy} as follows:

\begin{equation}
\mathcal{L}(y, p) = - \sum_{i=1}^{N} w_k \cdot y_i \cdot \log(p_i)
\end{equation}

where \( w_k \) represents the weight assigned to the language of the sentence \( k \). For each batch, the weight vector \( \mathbf{w}\) consists of elements with a value of 1 for all high-resource languages, and a value grater or equal to 1 for the low-resource language. We compute the loss of the batch using the mean of the losses. This approach allows us to customise the impact of different languages on the model's learning dynamics, improving its ability to handle language imbalances or prioritise the low-resource language over others during training.

In our research, we have explored different strategies for applying weights within the cross-entropy calculation. After analysis, we have identified the two methods that provide the best results. 

\subsection{Linear Progressive Weighted Cross-Entropy}
\label{ssec:lp_wce}

In the first approach, the weight for the low-resource language is calculated by a linear progression as a function of the training step. For the sake of simplicity, we omit the reference to the language \( k\), since it is only calculated for Galician. We define the value of the weight \( w_t\) as follows:

\begin{equation}
w_t = \begin{cases}
    1, & \text{if } t < t_{\text{min}} \\
    \alpha_{\text{ini}} + (\alpha_{\text{fin}} - \alpha_{\text{ini}}) \cdot \frac{\displaystyle (t - t_{\text{min}})}{\displaystyle t_{\text{total}} - t_{\text{min}}}, & \text{otherwise}
\end{cases}
\end{equation}

where \( w_t \) represents the weight value for a specific training step, \( \alpha_{\text{ini}} \) is the initial weight value, \( \alpha_{\text{fin}} \) is the final weight value, and the fraction represents the progression factor, where \( t \) denotes the current training step, \( t_{\text{min}} \) is the minimum number of steps before applying the weight, and \( t_{\text{total}} \) is the total number of training steps.

\subsection{Dynamic Weight Adaptation for Cross-Entropy}
\label{ssec:da_wce}

In the second approach, the weight for the low-resource language is calculated dynamically by adapting it to the ratio between the average losses for the low-resource language and the average losses for the other languages in the batch. This calculation is expressed as:

\begin{equation}
w_t = \begin{cases} 1, &
\text{if} \quad \frac{\displaystyle \overline{\mathcal{L}}_{\text{low}}}{\displaystyle \overline{\mathcal{L}}_{\text{high}}} {\displaystyle \cdot \alpha} < 1 \\
\max(\alpha, \frac{\displaystyle \overline{\mathcal{L}}_{\text{low}}}{\displaystyle \overline{\mathcal{L}}_{\text{high}}}), & \text{otherwise} \end{cases}
\end{equation}

where \( \overline{\mathcal{L}}_{\text{low}} \) is the average loss of the low-resource language in the batch, \( \overline{\mathcal{L}}_{\text{high}} \) is the average loss of the high-resource languages in the batch, and \( \alpha \) is the additional weighting factor to be given to the loss of the low-resource language. In this way, the dynamic weighting factor \( w_t \) is adjusted based on the ratio of the average losses in the batch multiplied by a factor \( \alpha \) at each step \( t \).

\section{Experimental Setup}
\label{sec:exp_setup}

\subsection{Description of the Dataset}

For the training and validation datasets we have used the Common Voice \cite{ardila2020common} splits in all languages. In the case of Galician, we also added the Common Voice test split to the training dataset. We balanced the data in each language so that they all had the same number of sentences, matching the language with the fewest hours available, in our case, Galician (Common Voice version 13.0 for Galician). For each language, we used 12 hours of audio for training and 7 hours of audio for validation.

For the high-resource languages test datasets, we have prepared splits from the Common Voice test sets (1.5 hours per language), but for the low-resource language, we have prepared a test split using FalAI \cite{pineiro2024falai}, a cross-domain dataset (also 1.5 hours). The aim is to test the robustness of the multilingual model for a domain other than the training domain, and against a larger and more diverse dataset.

\subsection{Data Augmentation Techniques}
\label{ssec:data_aug}

Previous research has consistently demonstrated the benefits of data augmentation, particularly in the domain of speech-related tasks \cite{ragni2014data}. Techniques such as time-stretching, gain adjustment, pitch shifting and noise addition have proven effective in keeping model performance robust in noisy environments and generalising well across different speakers \cite{park19e_interspeech}. 
We have developed a data augmentation pipeline that integrates these established techniques using the Python library audiomentations\footnote{\urlstyle{same}\url{https://github.com/iver56/audiomentations}}.

Applying these techniques, we aim to assess the impact of combining weighted cross-entropy with simple and proven augmentation methods applied only to the weighted class, i.\,e., the low-resource language. Our focus lies not in identifying the optimal augmentation techniques, but rather in assessing their impact when used in conjunction with weighted cross-entropy. By increasing the number of samples of the low-resource language, we will also increase the number of times that weighted cross-entropy is applied, as well as expanding the model's exposure to diverse training data, thereby capturing a more wider range of speech variations.

Our augmentation strategy includes the following techniques, effectively doubling the dataset size for the low-resource language:

\begin{itemize}
    \item Time Stretching: This technique varies the playback speed of audio while maintaining its pitch, thereby simulating different speaking rates and speech durations.
    \item Gain Adjustment: We introduce controlled changes in audio amplitude to replicate variations in recording conditions, volume levels, and microphone settings.
    \item Pitch Shifting: By altering the pitch of the audio, we mimic the natural variations in speakers' pitch, expanding the model's adaptability to different vocal characteristics.
    \item Gaussian Noise Addition: The introduction of Gaussian noise emulates background sounds and environmental variations, enhancing the model's resistance to real-world noise.
\end{itemize}

\subsection{Whisper Architecture and Fine-Tuning}

The Whisper checkpoints are available in five configurations of different model sizes. For our study, we fine-tune the multilingual version of the small checkpoint \emph{Whisper for Conditional Generation} with 244M parameters, equipped with 12 encoder and decoder layers, where each layer includes attention heads and feed-forward dimensions.

We perform a single multilingual and multitask fine-tuning. To do this, we use the tokenizer to encode the appropriate language label for each sample in the multilingual dataset and add it as part of the labels as a special token. In this way, we can perform a single fine-tuning, and although it is not the subject of this study, by introducing the language label, the model also learns to detect the spoken language with high accuracy as part of the transcription task. Figure~\ref{fig:sentence_w_special} shows an example of a sentence decoded with the language special token:

\begin{figure}[htbp]
  \centering
  \includegraphics[width=\linewidth]{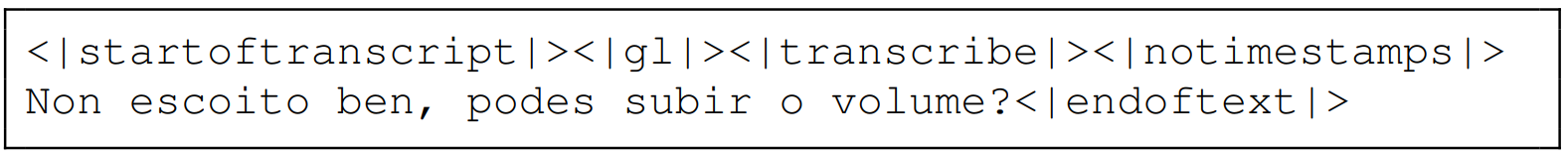}
  \caption{Sentence decoded with special tokens for Galician.}
  \label{fig:sentence_w_special}
\end{figure}

We use a 24GB NVIDIA Quadro RTX 6000 to fine-tune Whisper's small checkpoint, using the same maximum number of steps (8,000) and the same basic configuration in terms of batch size (16), learning rate ($10^{-5}$), gradient accumulation steps (1) or evaluation steps (1,000).
It should be noted that all models presented were trained from the same version of Whisper checkpoints, i.\,e., the starting point was the same for all fine-tuning procedures.

\begin{table*}[ht]
\centering
\captionsetup{justification=centering}
\caption{Word Error Rate (\%) of the multilingual ASR models obtained with our different approaches tested on the Common Voice dataset for Spanish, Portuguese, French, German and English; and on the in-house cross-domain dataset for Galician.}
\label{tab:asr_results}
\begin{tabular}{cccccccc} \toprule
\textbf{Model} & \textbf{Galician} & \textbf{Spanish} & \textbf{Portuguese} & \textbf{French} & \textbf{German} & \textbf{English} & \textbf{Mean} \\
\midrule
WS & 41.20 & 10.24 & 12.79 & 30.36 & 11.63 & 8.81 & 19.17 \\
WS-FT & 22.58 & 9.13 & 10.61 & 16.43 & 11.02 & 10.50 & 13.38 \\
WS-FT-LP-WCE & 21.43 & 10.00 & 10.87 & 16.61 & 11.19 & 10.31 & 13.40 \\
WS-FT$_{GL+}$ & 21.75 & 9.18 & 10.44 & 15.88 & 10.56 & 10.28 & 13.02 \\
WS-FT-LP-WCE$_{GL+}$ & \textbf{21.07} & 9.05 & 10.41 & 16.15 & 11.21 & 10.87 & 13.13 \\
WS-FT-DA-WCE$_{GL+}$ & 21.58 & 9.17 & 10.23 & 15.85 & 10.62 & 10.13 & \textbf{12.94} \\ \bottomrule
\end{tabular}

\end{table*}

\section{Results}
\label{sec:results}

\subsection{Baselines}
\label{ssec:baselines}

We base our investigation on the small checkpoint of the pre-trained Whisper model (\textbf{WS}). We evaluate our proposed approaches on five high-resource languages -- Spanish, Portuguese, French, German and English -- and one low-resource language -- Galician. The original Whisper model, trained on 96 languages, includes Galician in its corpus. However, due to limited data representation, recognition results for Galician are suboptimal without fine-tuning.

We explore the fine-tuned model (\textbf{WS-FT}) using balanced data across all six languages. In addition, as discussed in Section~\ref{ssec:data_aug}, we apply data augmentation techniques to address the challenges of low-resource languages by exploring the impact of increasing the number of training samples on the weighted cross-entropy (\textbf{WS-FT$_{GL+}$}).

In terms of weighted cross-entropy, our approaches are defined as follows: The first two use the linear progression method described in Section~\ref{ssec:lp_wce} to calculate the weights. The \textbf{WS-FT-LP-WCE} approach uses a linear progression with \( \alpha_{\text{ini}} \)=4, \( \alpha_{\text{fin}} \)=5, \( t_{\text{min}} \)=4k steps and \( t_{\text{max}} \)=8k steps without data augmentation, while the \textbf{WS-FT-LP-WCE$_{GL+}$} approach includes augmented data for Galician and uses a linear progression with \( \alpha_{\text{ini}} \)=2, \( \alpha_{\text{fin}} \)=5, \( t_{\text{min}} \)=4k steps and \( t_{\text{max}} \)=8k steps. For our third approach, we integrate the cross-entropy dynamic weight adaptation outlined in Section~\ref{ssec:da_wce}. In this scenario (\textbf{WS-FT-DA-WCE$_{GL+}$}), we set \( \alpha \)=1.5 and we include augmented data for Galician.

We experimentally determine the specific values of the parameters for all three approaches. Too high \( \alpha \) and \( \alpha_{\text{fin}} \) values lead to performance degradation in the other languages, while too low \( \alpha_{\text{ini}} \) values render the weighted cross-entropy ineffective in our first approach.

\subsection{Results and Discussion}

Table~\ref{tab:asr_results} shows the WER results for each of the baselines presented previously. The \textbf{WS} results correspond to the original Whisper model, without any fine-tuning. As expected, the results for the low-resource language are poor. The \textbf{WS-FT} show the results for the Whisper fine-tuned model using the same amount of data for each language. The fine-tuned model obtains better results in all languages except English, improving notably in the case of Galician and French. The degradation for the English results is consistent with similar Whisper fine-tuning studies \cite{rouditchenko23_interspeech}.

The direct application of the weighted cross-entropy without data augmentation (\textbf{WS-FT-LP-WCE}) manages to improve the results for Galician, confirming the effectiveness of the technique. However, such an aggressive application leads to a degradation in the results for the high-resource languages.

The results obtained by applying augmented data for Galician improve in this language, reducing the WER to 21.75\%. Applying this technique to Galician data also improves the recognition results in the rest of the languages thanks to the cross-lingual transfer. The fact of introducing more data (even if not in the target language) and making the model more robust to noise or gain improves the results in general.

Weighted cross-entropy combined with data augmentation further improves the results obtained by these techniques independently. Both weighting strategies improve the results for the low-resource language. The most aggressive strategy, \textbf{WS-FT-LP-WCE$_{GL+}$}, manages to reduce the WER for Galician the most, obtaining 21.07\%, at the cost of starting to degrade the other languages. On the other hand, the strategy of dynamic weight adaptation (\textbf{WS-FT-DA-WCE$_{GL+}$}) manages to obtain the most efficient result, reducing the WER for Galician to 21.58\%, and in turn reducing the WER for all languages except English.

In terms of relative reduction compared to the fine-tuned model without data augmentation, Table~\ref{tab:reduction} shows the results for each of our strategies. It demonstrates that we have been able to improve the recognition results for Galician, without degrading and even improving the results in the rest of the languages thanks to cross-lingual transfer properties.

The combination of weighted cross-entropy and data augmentation proves to be a particularly effective technique, capable of improving performance for the target language and underscoring that the utilization of weighted cross-entropy requires a larger number of samples on which to apply the weights.

\begin{table}[htbp]
\centering
\captionsetup{justification=centering}
\caption{Comparison of WER reductions relative to the fine-tuned model (\textbf{WS-FT}) for Galician and for the average of all languages.}
\label{tab:reduction}
\begin{tabular}{cC{2cm}C{2cm}} \toprule
\multirow{2}{*}{\textbf{Model}} & \textbf{Galician Reduction} & \textbf{Average Reduction} \\
\midrule
WS-FT & 0\% & 0\%  \\
WS-FT-LP-WCE & 5.09\% & -0.15\% \\
WS-FT$_{GL+}$ & 3.67\% & 2.69\%  \\
WS-FT-LP-WCE$_{GL+}$ & \textbf{6.69\%} & 1.87\%  \\
WS-FT-DA-WCE$_{GL+}$ & 4.43\% & \textbf{3.29\%} \\ \bottomrule
\end{tabular}

\end{table}

In conclusion, the use of weighted cross-entropy with dynamically adapted weights, guided by losses at each training step, combined with augmented data (\textbf{WS-FT-DA-WCE$_{GL+}$}), emerges as the most efficient strategy for achieving improved overall recognition results. This approach not only significantly improves results for low-resource languages, but also leverages cross-lingual transfer properties to improve recognition in four of the high-resource languages, proving to be robust and effective across different language contexts.

\section{Conclusions}
\label{sec:conclusions}

This work explores the integration of low-resource languages into multilingual ASR. We propose the use of language-weighted cross-entropy and its combination with data augmentation to improve results in the low-resource language without degrading results in the other languages while fine-tuning.
In conclusion, our approaches offer promising avenues for improving multilingual ASR systems for low-resource languages. Future work could focus on refining the strategies, exploring additional cross-lingual transfer properties, and optimising the adaptation of weights within this framework.

\section{Acknowledgements}
\label{ssec:acknow}

Work conducted as part of an internship at DFKI SLT Lab in Berlin from June to October 2023 by the first author. This research has been supported by the Galician Innovation Agency through the program ``Doutoramento Industrial'' and by the Xunta de Galicia through the grants ``Centro singular de investigación de Galicia accreditation 2019-2022'' and GPC ED431B 2021/24.

\bibliographystyle{IEEEtran}
\bibliography{template}

\begin{thebibliography}{10}
\providecommand{\url}[1]{#1}
\csname url@samestyle\endcsname
\providecommand{\newblock}{\relax}
\providecommand{\bibinfo}[2]{#2}
\providecommand{\BIBentrySTDinterwordspacing}{\spaceskip=0pt\relax}
\providecommand{\BIBentryALTinterwordstretchfactor}{4}
\providecommand{\BIBentryALTinterwordspacing}{\spaceskip=\fontdimen2\font plus
\BIBentryALTinterwordstretchfactor\fontdimen3\font minus \fontdimen4\font\relax}
\providecommand{\BIBforeignlanguage}[2]{{%
\expandafter\ifx\csname l@#1\endcsname\relax
\typeout{** WARNING: IEEEtran.bst: No hyphenation pattern has been}%
\typeout{** loaded for the language `#1'. Using the pattern for}%
\typeout{** the default language instead.}%
\else
\language=\csname l@#1\endcsname
\fi
#2}}
\providecommand{\BIBdecl}{\relax}
\BIBdecl

\bibitem{baevski2020effectiveness}
A.~Baevski and A.~Mohamed, ``Effectiveness of self-supervised pre-training for asr,'' in \emph{ICASSP 2020-2020 IEEE International Conference on Acoustics, Speech and Signal Processing (ICASSP)}.\hskip 1em plus 0.5em minus 0.4em\relax IEEE, 2020, pp. 7694--7698.

\bibitem{zhao2022improving}
J.~Zhao and W.-Q. Zhang, ``Improving automatic speech recognition performance for low-resource languages with self-supervised models,'' \emph{IEEE Journal of Selected Topics in Signal Processing}, vol.~16, no.~6, pp. 1227--1241, 2022.

\bibitem{miao2022multilingual}
L.~Miao, J.~Wu, P.~Behre, S.~Chang, and S.~Parthasarathy, ``Multilingual transformer language model for speech recognition in low-resource languages,'' in \emph{2022 Ninth International Conference on Social Networks Analysis, Management and Security (SNAMS)}.\hskip 1em plus 0.5em minus 0.4em\relax IEEE, 2022, pp. 1--5.

\bibitem{hadsell2020embracing}
R.~Hadsell, D.~Rao, A.~A. Rusu, and R.~Pascanu, ``Embracing change: Continual learning in deep neural networks,'' \emph{Trends in cognitive sciences}, vol.~24, no.~12, pp. 1028--1040, 2020.

\bibitem{kirkpatrick2017overcoming}
J.~Kirkpatrick, R.~Pascanu, N.~Rabinowitz, J.~Veness, G.~Desjardins, A.~A. Rusu, K.~Milan, J.~Quan, T.~Ramalho, A.~Grabska-Barwinska \emph{et~al.}, ``Overcoming catastrophic forgetting in neural networks,'' \emph{Proceedings of the national academy of sciences}, vol. 114, no.~13, pp. 3521--3526, 2017.

\bibitem{wang2015transfer}
D.~Wang and T.~F. Zheng, ``Transfer learning for speech and language processing,'' in \emph{2015 Asia-Pacific Signal and Information Processing Association Annual Summit and Conference (APSIPA)}.\hskip 1em plus 0.5em minus 0.4em\relax IEEE, 2015, pp. 1225--1237.

\bibitem{radford2023robust}
A.~Radford, J.~W. Kim, T.~Xu, G.~Brockman, C.~McLeavey, and I.~Sutskever, ``Robust speech recognition via large-scale weak supervision,'' in \emph{International Conference on Machine Learning}.\hskip 1em plus 0.5em minus 0.4em\relax PMLR, 2023, pp. 28\,492--28\,518.

\bibitem{ardila2020common}
R.~Ardila, M.~Branson, K.~Davis, M.~Kohler, J.~Meyer, M.~Henretty, R.~Morais, L.~Saunders, F.~Tyers, and G.~Weber, ``Common voice: A massively-multilingual speech corpus,'' in \emph{Proceedings of the Twelfth Language Resources and Evaluation Conference}, 2020, pp. 4218--4222.

\bibitem{pineiro2024falai}
A.~Pi{\~n}eiro-Mart{\'\i}n, C.~G. Mateo, L.~Doc{\'\i}o-Fern{\'a}ndez, M.~del Carmen L{\'o}pez-P{\'e}rez, and J.~Gandarela-Rodr{\'\i}guez, ``Falai: A dataset for end-to-end spoken language understanding in a low-resource scenario,'' in \emph{Proceedings of the 2024 Joint International Conference on Computational Linguistics, Language Resources and Evaluation (LREC-COLING 2024)}, 2024, pp. 7107--7116.

\bibitem{baevski2019vq}
A.~Baevski, S.~Schneider, and M.~Auli, ``vq-wav2vec: Self-supervised learning of discrete speech representations,'' in \emph{International Conference on Learning Representations}, 2019.

\bibitem{babu22_interspeech}
A.~Babu, C.~Wang, A.~Tjandra, K.~Lakhotia, Q.~Xu, N.~Goyal, K.~Singh, P.~{von Platen}, Y.~Saraf, J.~Pino, A.~Baevski, A.~Conneau, and M.~Auli, ``{XLS-R: Self-supervised Cross-lingual Speech Representation Learning at Scale},'' in \emph{Proc. Interspeech 2022}, 2022, pp. 2278--2282.

\bibitem{zhang2023google}
Y.~Zhang, W.~Han, J.~Qin, Y.~Wang, A.~Bapna, Z.~Chen, N.~Chen, B.~Li, V.~Axelrod, G.~Wang \emph{et~al.}, ``Google usm: Scaling automatic speech recognition beyond 100 languages,'' \emph{arXiv preprint arXiv:2303.01037}, 2023.

\bibitem{pratap2023scaling}
V.~Pratap, A.~Tjandra, B.~Shi, P.~Tomasello, A.~Babu, S.~Kundu, A.~Elkahky, Z.~Ni, A.~Vyas, M.~Fazel{-}Zarandi, A.~Baevski, Y.~Adi, X.~Zhang, W.~Hsu, A.~Conneau, and M.~Auli, ``Scaling speech technology to 1,000+ languages,'' \emph{CoRR}, vol. abs/2305.13516, 2023.

\bibitem{barrault2023seamless}
L.~Barrault, Y.-A. Chung, M.~C. Meglioli, D.~Dale, N.~Dong, M.~Duppenthaler, P.-A. Duquenne, B.~Ellis, H.~Elsahar, J.~Haaheim \emph{et~al.}, ``Seamless: Multilingual expressive and streaming speech translation,'' \emph{arXiv preprint arXiv:2312.05187}, 2023.

\bibitem{narayanan2018toward}
A.~Narayanan, A.~Misra, K.~C. Sim, G.~Pundak, A.~Tripathi, M.~Elfeky, P.~Haghani, T.~Strohman, and M.~Bacchiani, ``Toward domain-invariant speech recognition via large scale training,'' in \emph{2018 IEEE Spoken Language Technology Workshop (SLT)}.\hskip 1em plus 0.5em minus 0.4em\relax IEEE, 2018, pp. 441--447.

\bibitem{likhomanenko21_interspeech}
T.~Likhomanenko, Q.~Xu, V.~Pratap, P.~Tomasello, J.~Kahn, G.~Avidov, R.~Collobert, and G.~Synnaeve, ``{Rethinking Evaluation in ASR: Are Our Models Robust Enough?}'' in \emph{Proc. Interspeech 2021}, 2021, pp. 311--315.

\bibitem{khare2021low}
S.~Khare, A.~R. Mittal, A.~Diwan, S.~Sarawagi, P.~Jyothi, and S.~Bharadwaj, ``Low resource asr: The surprising effectiveness of high resource transliteration.'' in \emph{Interspeech}, 2021, pp. 1529--1533.

\bibitem{meng2021mixspeech}
L.~Meng, J.~Xu, X.~Tan, J.~Wang, T.~Qin, and B.~Xu, ``Mixspeech: Data augmentation for low-resource automatic speech recognition,'' in \emph{ICASSP 2021-2021 IEEE International Conference on Acoustics, Speech and Signal Processing (ICASSP)}.\hskip 1em plus 0.5em minus 0.4em\relax IEEE, 2021, pp. 7008--7012.

\bibitem{ragni2014data}
A.~Ragni, K.~M. Knill, S.~P. Rath, and M.~J. Gales, ``Data augmentation for low resource languages,'' in \emph{INTERSPEECH 2014: 15th Annual Conference of the International Speech Communication Association}.\hskip 1em plus 0.5em minus 0.4em\relax International Speech Communication Association (ISCA), 2014, pp. 810--814.

\bibitem{aurelio2019learning}
Y.~S. Aurelio, G.~M. De~Almeida, C.~L. de~Castro, and A.~P. Braga, ``Learning from imbalanced data sets with weighted cross-entropy function,'' \emph{Neural processing letters}, vol.~50, pp. 1937--1949, 2019.

\bibitem{chen2017learning}
G.~Chen, W.~Choi, X.~Yu, T.~Han, and M.~Chandraker, ``Learning efficient object detection models with knowledge distillation,'' \emph{Advances in neural information processing systems}, vol.~30, 2017.

\bibitem{panchapagesan16_interspeech}
S.~Panchapagesan, M.~Sun, A.~Khare, S.~Matsoukas, A.~Mandal, B.~Hoffmeister, and S.~Vitaladevuni, ``{Multi-Task Learning and Weighted Cross-Entropy for DNN-Based Keyword Spotting},'' in \emph{Proc. Interspeech 2016}, 2016, pp. 760--764.

\bibitem{Gaspari2023}
\BIBentryALTinterwordspacing
F.~Gaspari, A.~Gr{\"u}tzner-Zahn, G.~Rehm, O.~Gallagher, M.~Giagkou, S.~Piperidis, and A.~Way, \emph{Digital Language Equality: Definition, Metric, Dashboard}.\hskip 1em plus 0.5em minus 0.4em\relax Cham: Springer International Publishing, 2023, pp. 39--73. [Online]. Available: \url{https://doi.org/10.1007/978-3-031-28819-7_3}
\BIBentrySTDinterwordspacing

\bibitem{park19e_interspeech}
D.~S. Park, W.~Chan, Y.~Zhang, C.-C. Chiu, B.~Zoph, E.~D. Cubuk, and Q.~V. Le, ``{SpecAugment: A Simple Data Augmentation Method for Automatic Speech Recognition},'' in \emph{Proc. Interspeech 2019}, 2019, pp. 2613--2617.

\bibitem{rouditchenko23_interspeech}
A.~Rouditchenko, S.~Khurana, S.~Thomas, R.~Feris, L.~Karlinsky, H.~Kuehne, D.~Harwath, B.~Kingsbury, and J.~Glass, ``{Comparison of Multilingual Self-Supervised and Weakly-Supervised Speech Pre-Training for Adaptation to Unseen Languages},'' in \emph{Proc. INTERSPEECH 2023}, 2023, pp. 2268--2272.

\end{thebibliography}

\end{document}